# IR-LPR: Large Scale of Iranian License Plate Recognition Dataset

Mahdi Rahmani
Faculty of Electrical & Computer Engineering
Malek Ashtar University of Technology, Iran
mahdirahmani1375@gmail.com

Melika Sabaghian
Faculty of Electrical & Computer Engineering
Malek Ashtar University of Technology, Iran
melika.sabaghian@gmail.com

Seyyede Mahila Moghadami
Faculty of Electrical & Computer Engineering
Malek Ashtar University of Technology, Iran
mahilamoghadami@gmail.com

Mohammad Mohsen Talaie
Faculty of Electrical & Computer Engineering
Malek Ashtar University of Technology, Iran
m.m.talaie@live.com

Mahdi Naghibi
Faculty of Electrical & Computer Engineering
Malek Ashtar University of Technology, Iran
mahdi110@gmail.com

Mohammad Ali Keyvanrad
Faculty of Electrical & Computer Engineering
Malek Ashtar University of Technology, Iran
keyvanrad@mut.ac.ir

*Abstract*— Object detection has always been practical. There are so many things in our world that recognizing them can not only increase our automatic knowledge of the surroundings, but can also be lucrative for those interested in starting a new business. One of these attractive objects is the license plate (LP). In addition to the security uses that license plate detection can have, it can also be used to create creative businesses. With the development of object detection methods based on deep learning models, an appropriate and comprehensive dataset becomes doubly important. But due to the frequent commercial use of license plate datasets, there are limited datasets not only in Iran but also in the world. The largest Iranian dataset for detection license plates has 1,466 images. Also, the largest Iranian dataset for recognizing the characters of a license plate has 5,000 images. We have prepared a complete dataset including *20,967* car images along with all the detection annotation of the whole license plate and its characters, which can be useful for various purposes. Also, the total number of license plate images for character recognition application is *27,745* images.

Download: https://github.com/mut-deep/IR-LPR

## I. Introduction

### A. Automatic License Plate Recognition(ALPR)

In recent years automatic license plate recognition (ALPR) also known as automatic number plate recognition (ANPR) has been a frequent topic of research[1], [2] due to many practical applications. ALPR refers to the process of using artificial intelligence methods to recognize the characters in the vehicles' license plates. ALPR system which can be found in many recent approaches [3]–[6] consist of three main steps, License Plate Detection (LPD), Character Segmentation (CS), and Optical Character Recognition (OCR). there are still some challenges in building an efficient ALPR system such as occlusion, various license plate layouts and languages, noisy or dirty input images, different sizes and width-to-height ratios of the license plates, and a variety of illumination and weather conditions[1]. Employing an appropriate method for each of the mentioned steps to handle these challenges with the lowest possible side effects is an essential task.

### B. ALPR Challenges

In this section we study challenges which an ALPR problem face with:

*1) Different weather conditions:* Both in plate detection and character recognition task, extreme weather conditions such as rainy, snow, fog and other similar conditions, can make challenges and difficulties[7].

*2) Light radiation angle:* Direct and stunning sunshine, formated other vehicles and objects shadow in effect of different light radiation angle are problems which accured during the day. Furthermore, direct radiated light from other objects and especially vehicles during night, can be another challenge which face with.

*3) Different distortions:* Although objects with frontal view still have challenges in different dimensions such as distance, but variety of different distortions make detection task more challengeable[8].

*4) Similarity of individual characters:* In task of character recognition, similarity between characters such as "س" and "ص", and, "۲" and "۳" in persian plate dataset, is another important point which need to be handled.

*5) Tiny and new characters:* In persian plate dataset, " 0 " is a new and also tiny character which recently assigned to plate labels. So because of its lack of number, need more attention in augmenting procedure[6].

*6) Special plates:* Low number of special plates such as governmental, temporary and etc versus private plates make model weaker against these special types.

### C. Why LP Dataset is Important?

One of the main challenges of having a good ALPR system is having a good dataset, a dataset that not only has a large number of images but also includes images with different angles, distances, under different illumination (day and night) and different conditions[1]. Unfortunately, there are not many datasets for Iranian license plates. The dataset introduced in this article can be a good one for many academic or business purposes such as vehicle identification, automatic toll collection, stolen vehicles detection, traffic law enforcement, private spaces access control, road traffic monitoring, etc.

In the first part, an introduction to the dataset was provided. In the second part, we will fully examine our own dataset and compare it with other datasets. Due to the low number of some types of license plates, in addition to the real dataset, we have also created a dummy dataset, the details of which are given in this part. In the third part, several models have been trained on our dataset and finally its performance has been examined. And at the end, a conclusion of the work done is presented.

## II. IR-LPR Dataset Description

### A. Iranian LP Format

There is a standard format for vehicles license plates in Iran, Iranian license plates consist of two parts, the leftmost part presents a unique string of numbers and characters for the vehicle, which includes two digits on the left, an alphabetic character in the middle, and three digits on the right. Range of digits in this part is from 0 to 9 and characters for the middle part are selected from 26 alphabet letters (Includes 8 extremely rare letters among the selected letters), the rightmost part consist of a two-digit code (with reduced font size), which refers to the city where the vehicle owner lives. Besides, license plates background colors may vary regarding the type and usage of the vehicle. For instance, private cars have a white background while public vehicles and Taxis have a yellow background. some samples of standard format for vehicles license plates in Iran represented in Figure 1.

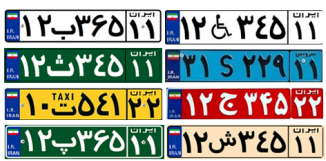

*Figure 1 Some samples of standard format for Iranian LPs*

As it's shown in Figure 1 private cars for people with disabilities have a wheelchair symbol instead of the Farsi character. additionally, free trade zone license plates have a different width-to-height ratio, this kind of license plate consists of two parts, the leftmost part is blue and holds the free trade zone's logo, i.e., Anzali, Kish, Arvand, Aras etc. the rightmost part is white and contains just five numbers alongside their identical English[9].

### B. Our Real Iranian LP Dataset

The IR-LPR dataset has been collected from 2018 to 2022. This dataset has been photographed and edited by at least 55 machine learning students at Malek Ashtar University of Technology. Most students used their cell phone camera to take pictures.

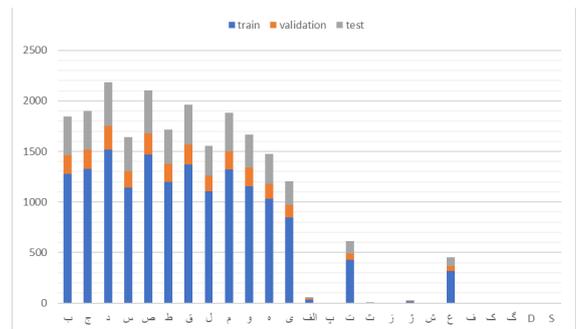

*Figure 2 The distribution of LP types in the our dataset*

The resolution of some images was too high, which in addition to large volume, was not suitable for use in models. For this reason, very large images have been resized to a maximum height of 1280 pixels and a maximum width of 1280 pixels.

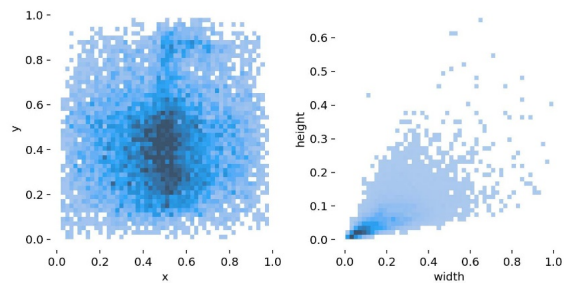

*Figure 3 LP location and size*

Students before use LabelImg tool, used our own programming software for labeling. But then the LabelImg tool was used for this purpose.

The IR-LPR dataset contains a total of 19,937 images. The Table 1 shows the details of this dataset based on night/day and train/validation/test.

*Table 1 Details of our dataset*

|  | Train | validation | test | **total** |
|---|---|---|---|---|
| day | 11,786 | 1,708 | 3,351 | 16,845 |
| night | 2,885 | 412 | 825 | 4,122 |
| **total** | **14,671** | **2,120** | **4,176** | **20,967** |

In addition to the numbers, 26 other symbols are used on the Iranian LP. The Table 2 shows the number of each symbol in our dataset.

*Table 2 The number of each symbol in our dataset*

|  |  |  | train | validation | test | total |
|---|---|---|---|---|---|---|
|  | LPs |  | 19,381 | 2,805 | 5,559 | 27,745 |
| Private Vehicles | B | ب | 1,491 | 216 | 432 | 2,139 |
|  | J | ج | 1,480 | 214 | 424 | 2,118 |
|  | D | د | 1,785 | 256 | 513 | 2,554 |
|  | S | س | 1,372 | 199 | 395 | 1,966 |
|  | Ṣ | ص | 1,723 | 248 | 491 | 2,462 |
|  | Ṭ | ط | 1,355 | 197 | 390 | 1,942 |
|  | Q | ق | 1,579 | 228 | 453 | 2,260 |
|  | L | ل | 1,214 | 177 | 347 | 1,738 |
|  | M | م | 1,529 | 217 | 435 | 2,181 |
|  | V | و | 1,430 | 206 | 410 | 2,046 |
|  | H | ه | 1,143 | 163 | 326 | 1,632 |
|  | N | ن | 1,145 | 167 | 327 | 1,639 |
|  | Y | ی | 1,056 | 153 | 302 | 1,511 |
| Governmental Vehicles | A | الف | 54 | 8 | 16 | 78 |
| Police Vehicles | P | پ | 1 | 1 | - | 2 |
| Taxis | T | ت | 645 | 94 | 185 | 924 |
| IRGC Vehicles | Ṯ | ث | 1 | 1 | 1 | 3 |
| Ministry of Defence | Z | ز | 1 | - | - | - |
| Private vehicles of people with disabilities | Ž | ژ (ج) | 23 | 4 | 7 | 34 |
| Islamic Republic of Iran Army Vehicles | Š | ش | 1 | 1 | - | 2 |
| Public Vehicles | O | ع | 421 | 61 | 119 | 601 |
| General Staff of Armed Forces Vehicles | F | ف | - | - | - | - |
| Agricultural Vehicles | K | ک | - | - | - | - |
| Temporary Passage | G | گ | - | - | - | - |
| Diplomatic |  | D | - | - | - | - |
| Consular and International Services |  | S | - | - | - | - |
| Numbers | 0 | ۰ | 2,014 | 296 | 555 | 2,865 |
|  | 1 | ۱ | 17,928 | 2,623 | 5,080 | 25,633 |
|  | 2 | ۲ | 14,852 | 2,105 | 4,260 | 21,221 |
|  | 3 | ۳ | 13,308 | 1,852 | 3,791 | 18,957 |
|  | 4 | ۴ | 14,018 | 2,006 | 4,087 | 20,119 |
|  | 5 | ۵ | 12,820 | 1,857 | 3,633 | 18,320 |
|  | 6 | ۶ | 14,544 | 2,104 | 4,184 | 20,844 |
|  | 7 | ۷ | 13,177 | 1,944 | 3,846 | 18,981 |
|  | 8 | ۸ | 14,938 | 2,160 | 4,328 | 21,442 |
|  | 9 | ۹ | 14,376 | 2,116 | 4,112 | 20,622 |

Also, in the Figure 2, the distribution of LP types in the IR-LPR dataset can be seen.

As it turns out, our dataset is complete for general use. For cases where all types of LPs need to be trained, either a new photo must be taken or a dummy dataset must be created. To solve this problem, we have generated dummy data in the next section. In terms of LP location and size, the Figure 3 can be drawn.

As shown in the Figure 3, the scatter of LPs in the images is often in the middle. However, in other locations there were also significant numbers of LPs. LP size scatter also indicates that most LPs have a standard size for real-world detection.

*Table 3 Our dataset vs other datasets*

| Name | Type | Number | Country | Condition |
|---|---|---|---|---|
| IRCP dataset[12] | Car | 220 | Iran | various illumination conditions and distance |
| Car license plate dataset | Car | 1,466 | Iran | various illumination conditions and distance |
| A Large-scale Dataset of Farsi License Plate Characters[9] | Character | 83,844 | Iran | real-world captured by various cameras |
| IRVD[13] | Vehicle | - | Iran | Rainy, sunny, cloudy, night |
| AOLP[14] | LP | 2,049 | Taiwan | different weather + distance |
| AOLPE[7] | LP | 4,200 | Taiwan | (Extreme Weather, Scene Complexity, Glaring / Lighting Effects |
| Cars Dataset (Stanford university)[15] | Car | 16,185 | USA | - |
| Sense SegPlate Database[16] | LPCS | 2,107 | Brazil | - |
| Sense-ALPR Database[17] | Car | 8,683 | Brazil | - |
| Paper (Wuhan university)[18] | Car | 7,284 | China | different orientations and multiple scales |
| Paper (Wuhan university)[18] | LP | 10,279 | China | different orientations and multiple scales |
| Our dataset | Car (annotated) | 20,967 | Iran | Day/night, different orientations, multiple scales, Lighting Effects, captured by various cameras, different weather |
| Our dataset | LP (annotated) | 27,745 | Iran | Day/night, different orientations, multiple scales, Lighting Effects, captured by various cameras, different weather |
| Our dataset | Just Dummy LP (annotated) | 28,600 | Iran | Includes different types of Iranian license plates with different types of noise |

In addition to the main dataset, we have prepared a dataset of LPs for LP recognition task. This dataset contains 27,867 LPs. You can see examples of this dataset in the Figure 4.

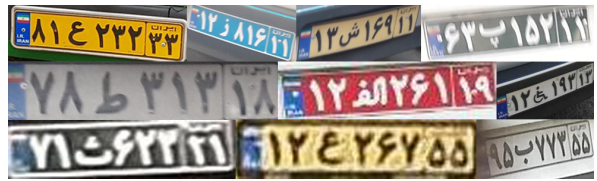

*Figure 4 Part of LP dataset*

### C. Our Dummy Iranian LP Dataset

As mentioned in the previous section, some types of LPs have not been photographed or have been photographed poorly. For this reason, the solution of producing dummy LP data was chosen to solve this problem. In the Figure 5 you can see part of this dataset. We used the two repositories [10], [11] to create an dummy Iranian LP dataset.

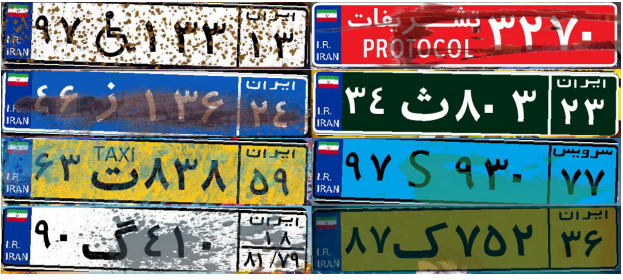

*Figure 5 Part of dummy dataset*

### D. Our Iranian LP Dataset vs Other Datasets

There are different datasets in Iran and the world. We have reviewed some of the largest of these datasets and compared them with our own dataset in terms of criteria of type, country, number, etc. The results can be seen in the Table 3.

## III. EVALUATE THE DATASET

We intend to evaluate several deep learning models to evaluate the dataset. Therefore, we will first explain the common pipeline in LP detection and recognition, and then go to the explanation of its parts.

### A. ALPR Pipeline

The pipeline in Figure 6 can be considered to increase the accuracy of detection and recognition.

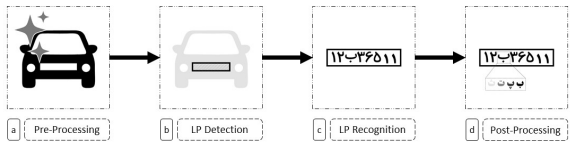

*Figure 6 ALPR pipeline*

a) Pre-Processing: This step itself can include two steps of generating various data for the training stage (data augmentation) and increasing the image quality using various image processing methods (for example, removing noise, increasing resolution, etc.) for the test stage.
b) LP Detection: A license plate (LP) is an object. Therefore, using the methods of object detection in the image, such as YOLO, faster R-CNN, SSD, Detectron2, etc., as a license plate detection operation can be performed.
c) LP Recognition: After detecting the LP, it is time to recognition (read) the LP. At this stage, two general methods of image processing (such as using histograms and classification) or deep learning (such as object detection) can be used.
d) Post-Processing: LP recognition and reading operation may not always be accurate. Low image quality, the presence of noise, etc. are the influential factors at this stage. For this reason, it is necessary to use additional information to increase reading accuracy (Because the LP has a specific format and not every number and symbol can be placed anywhere, we can reduce a large number of errors.).

In this paper, we have implemented steps a: various volumes of real images, b: YOLO and Detectron2 and c: YOLO and Detectron2.

### B. LP Detection

We used 2 methods for detect LP:

*1) YOLOv5[19]:* Yolo is a state of the art Object Detector which can perform object detection in real-time due to its speed and accuracy in which divides images into a grid system. YOLO v5 is different from all other prior releases, as this is a PyTorch implementation rather than a fork from original Darknet. The major improvements in yolov5 includes mosaic data augmentation and auto learning bounding box anchors[20].

*2) Detectron2:* "Detectron2 is Facebook AI Research's next-generation software system that implements state-of-the-art object detection algorithms[21]" Detectron2 includes a variety of models like Faster R-CNN, Mask R-CNN, RetinaNet, DensePose, Cascade R-CNN, Panoptic FPN, and TensorMask. It provides support for many different computer vision tasks including object detection, instance segmentation, human pose prediction, and panoptic segmentation. Detectron2 is built using Pytorch.

*Table 4 Detection evaluation*

| | Model | Precision | Recall | mAP@.5 | mAP@.5:.95 |
|---|---|---|---|---|---|
| LP | YOLOv5-s | 0.954 | 0.956 | 0.982 | 0.817 |
| | YOLOv5-x | 0.962 | 0.957 | 0.985 | 0.83 |
| | Detectron2 | 0.865 | 0.848 | 0.975 | 0.735 |
| Character | YOLOv5-s | 0.934 | 0.935 | 0.944 | 0.8 |
| | YOLOv5-x | 0.922 | 0.944 | 0.943 | 0.804 |
| | Detectron2 | 0.759 | 0.758 | 0.842 | 0.628 |
| | YOLOv5-s (+dummy dataset) | 0.941 | 0.936 | 0.939 | 0.793 |
| | YOLOv5-x (+dummy dataset) | 0.945 | 0.945 | 0.948 | 0.812 |
| | Detectron2 (+dummy dataset) | 0.927 | 0.903 | 0.964 | 0.859 |

### C. LP Recognition

We also used the same two YOLOv5 and Detectron2 methods for Recognition.

### D. Evaluation Metrics

After training the model, it is time to evaluate. We perform the license plate reading operation in two stages. For this reason, we use four evaluation criteria precision, recall, mAP@.5 and mAP@.5:.95 to evaluate the detection of LP from the background and also the detection of characters from the LP (Table 4).

Also, to evaluate the accuracy of the read license plate, the 100% accurate LP reading and Levenshtein ratio criteria are used (Table 5).

*Table 5 LP reading evaluation*

|  | Model | Accuracy |
|---|---|---|
| 100% accurate LP reading | YOLOv5-s | 0.86 |
|  | YOLOv5-x | 0.867 |
|  | Detectron2 | 0.913 |
|  | YOLOv5-s (+dummy dataset) | 0.852 |
|  | YOLOv5-x (+dummy dataset) | 0.862 |
|  | Detectron2 (+dummy dataset) | 0.933 |
| Levenshtein ratio | YOLOv5-s | 0.977 |
|  | YOLOv5-x | 0.979 |
|  | Detectron2 | 0.959 |
|  | YOLOv5-s (+dummy dataset) | 0.976 |
|  | YOLOv5-x (+dummy dataset) | 0.979 |
|  | Detectron2 (+dummy dataset) | 0.971 |

## IV. Conclusion

The number and variety of our datasets is so high that it can be used for many different purposes. We have prepared a complete dataset including 20,967 car images along with all the detection annotation of the whole license plate and its characters, which can be useful for various purposes. Also, the total number of license plate images for character recognition application is 27,745 images.

The trained models on this dataset, which are provided to you on GitHub, allow you to use this dataset more easily for your intended purposes.

## Acknowledgment

We thank all the students of Malek Ashtar University of Technology in Iran who have helped us in preparing this dataset.